# Direct Method for Training Feed-forward Neural Networks using Batch Extended Kalman Filter for Multi-Step-Ahead Predictions

Artem Chernodub,

Institute of Mathematical Machines and Systems NASU, Neurotechnologies Dept., Glushkova 42 ave., 03187 Kyiv, Ukraine

a.chernodub@gmail.com

This paper is dedicated to the long-term, or multi-step-ahead, time series prediction problem. We propose a novel method for training feed-forward neural networks, such as multilayer perceptrons, with tapped delay lines. Special batch calculation of derivatives called Forecasted Propagation Through Time and batch modification of the Extended Kalman Filter are introduced. Experiments were carried out on well-known timeseries benchmarks, the Mackey-Glass chaotic process and the Santa Fe Laser Data Series. Recurrent and feed-forward neural networks were evaluated.

**Keywords:** Forecasted Propagation Through Time, multi-step-ahead prediction, Batch Extended Kalman Filter

## 1   Introduction

Time series forecasting is a current scientific problem that has many applications in control theory, economics, medicine, physics and other domains. Neural networks are known as an effective and friendly tool for black-box modeling of unknown plant's dynamics [1]. Usually, neural networks are trained to perform single-step-ahead (SS) predictions, where the predictor uses some available input and output observations to estimate the variable of interest for the time step immediately following the latest observation [2-4]. However, recently there has been growing interest in multi-step-ahead (MS) predictions, where the values of interest must be predicted for some horizon in the future. Knowing the sequence of future values allows for estimation of projected amplitudes, frequencies, and variability, which are important for modeling predictive control [5], flood forecasts [6], fault diagnostics [7], and web server queuing systems [8]. Generally speaking, the ability to perform MS predictions is frequently treated as the "true" test for the quality of a developed empirical model. In particular, well-known echo state machine neural networks (ESNs) became popular because of their ability to perform good long-horizon ($H = 84$) multistep predictions.

The most straightforward approach to perform MS prediction is to train the SS predictor first and then use it in an autonomous "closed-loop" mode. The predictor's output is fed back to the input for a finite number of time steps. However, this simple method frequently shows poor results because of the accumulation of errors on difficult data points [4]. Recurrent neural networks (RNNs) such as NARX and Elman networks usually show better results. They are based on the calculation of special dynamic derivatives called Backpropagation Through Time (BPTT). The underlying idea of BPTT is to calculate derivatives by propagating the errors back across the RNN, which is unfolded through time. This penalizes the predictor for accumulating errors in time and therefore provides better MS predictions. Nonetheless, RNNs have some disadvantages. First, the implementation of RNNs is harder than feed-forward neural networks (FFNNs) in industrial settings. Second, training the RNNs is a difficult problem because of their more complicated error surfaces and vanishing gradient effects [9]. Third, the internal dynamics of RNNs make them less friendly for stability analysis. All of the above reasons prevent RNNs from becoming widely popular in industry. Meanwhile, RNNs have inspired a new family of methods for training FFNNs to perform MS predictions called direct methods [4]. Accumulated error is backpropagated through an unfolded through time FFNN in BPTT style that causes minimization of the MS prediction error. Nevertheless, the vanishing gradient effect still occurs in all multilayer perceptron-based networks with sigmoidal activation functions.

We propose a new, effective method for training the feed-forward neural models to perform MS prediction, called Forecasted Propagation Through Time (FPTT), for calculating the batch-like dynamic derivatives that minimize the negative effect of vanishing gradients. We use batch modification of the EKF algorithm which naturally deals with these batch-like dynamic derivatives for training the neural network.

## 2 Modeling time series dynamics

We consider modeling time series in the sense of dealing with generalized nonlinear autoregression (NAR) models. In this case, time series behavior can be captured by expressing the observable value $y(k+1)$ as a function of $N$ previous values $y(k),..., y(k-N+1)$:

$$y(k+1) = F(y(k), y(k-1),..., y(k-N+1)), \quad (1)$$

where $k$ is the time step variable and $F(\cdot)$ is an unknown function that defines the underlying dynamic process. The goal of training the neural network is to develop the empirical model of function $F(\cdot)$ as closely as possible. If such a neural model $\tilde{F}(\cdot)$ is available, one can perform iterated multi-step-ahead prediction:

$$\tilde{y}(k+1) = \tilde{F}(y(k), y(k-1),..., y(k-N+1)), \quad (2)$$

$$\ldots$$

$$\tilde{y}(k+H+1) = \tilde{F}(\tilde{y}(k+H), \tilde{y}(k+H-1),..., \tilde{y}(k+H-N+1)), \quad (3)$$

where $\tilde{y}$ is the neural network's output and $H$ is the horizon of prediction.

### 2.1 Training traditional Multilayer Perceptrons using EKF for SS predictions

**Dynamic Multilayer Perceptron.** Dynamic multilayer perceptrons (DMLP) are the most popular neural network architectures for time series prediction. Such neural networks consist of multilayer perceptrons with added tapped delay line of order $N$ (Fig. 1, left).

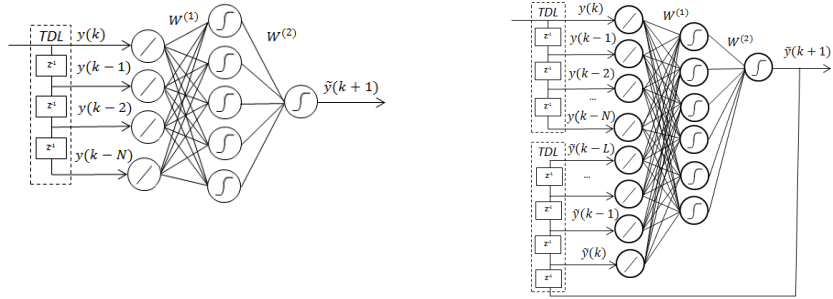

**Fig. 1.** DMLP neural network (left), NARX neural network (right).

The neural network receives an input vector $x(k) = [y(k) \ \ y(k-1) \ \ .... \ \ y(k-N)]^T$, and calculates the output $\tilde{y}(k+1) = g(\sum_j w_j^{(2)}(f(\sum_i w_{ji}^{(1)} x_i)))$, where $w^{(1)}$ and $w^{(2)}$ are weights of the hidden and output layers and $f(\cdot)$ and $g(\cdot)$ are activation functions of the hidden and output layers.

**Calculation of BP derivatives.** The Jacobians $\frac{\partial \tilde{y}}{\partial w}$ for the neural network's training procedure are calculated using a standard backpropagation technique by propagating a constant value $\delta^{OUT} = 1$ at each backward pass instead of propagating the residual error $\delta^{OUT} = t(k+1) - \tilde{y}(k+1)$ which calculates Jacobians $\frac{\partial \tilde{y}(k)}{\partial w}$ instead of error gradients $\frac{\partial E(k)}{\partial w}$ because $\frac{\partial E(k)}{\partial w} = \frac{\partial [e(k+1)^2]}{\partial w} = 2e(k+1)\frac{\partial y}{\partial w}$.

**Extended Kalman Filter method for training DMLP.** Although EKF training [7], [10] is usually associated with RNNs it can be applied to any differentiable parameterized model. Training the neural network using an Extended Kalman Filter may be considered a state estimation problem of some unknown "ideal" neural network that provides zero residual. In this case, the states are the neural network's weights $w(k)$ and the residual is the current training error $e(k+1) = t(k+1) - \tilde{y}(k+1)$. During the initialization step, covariance matrices of measurement noise $R = \eta I$ and dynamic training noise $Q = \mu I$ are set. Matrix $R$ has size $L_w \times L_w$, matrix $Q$ has size $N_w \times N_w$, where $L_w$ is the number of output neurons, and $N_w$ is the number of the network's weight coefficients. Coefficient $\eta$ is the training speed, usually $\eta \sim 10^{-2}...10^{-4}$, and coefficient $\mu$ defines the measurement noise, usually $\mu \sim 10^{-4}...10^{-8}$. Also, the identity covariance matrix $P$ of size $N_w \times N_w$ and zero observation matrix $H$ of size $L_w \times N_w$ are defined. The following steps must be performed for all elements of the training dataset:

1) Forward pass: the neural network's output $\tilde{y}(k+1)$ is calculated.

2) Backward pass: Jacobians $\frac{\partial \tilde{y}}{\partial w}$ are calculated using backpropagation. Observation matrix $H(k)$ is filled:

$$H(k) = \left[ \frac{\partial \tilde{y}(k+1)}{\partial w_1} \quad \frac{\partial \tilde{y}(k+1)}{\partial w_2} \quad ... \quad \frac{\partial \tilde{y}(k+1)}{\partial w_{N_w}} \right]. \quad (4)$$

3) Residual matrix $E(k)$ is filled:

$$E(k) = [e(k+1)]. \quad (5)$$

4) New weights $w(k)$ and correlation matrix $P(k+1)$ are calculated:

$$K(k) = P(k)H(k)^T [H(k)P(k)H(k)^T + R]^{-1}, \quad (6)$$

$$P(k+1) = P(k) - K(k)H(k)P(k) + Q, \quad (7)$$

$$w(k+1) = w(k) + K(k)E(k). \quad (8)$$

**2.2 Training NARX networks using BPTT and EKF**

**Nonlinear Autoregression with eXternal inputs.** The NARX neural network structure is shown in Fig. 1. It is equipped with both a tapped delay line at the input and global recurrent feedback connections, so the input vector $x(k) = [y(k) \quad ... \quad y(k-N) \quad \tilde{y}(k) \quad ... \quad \tilde{y}(k-L)]^T$, where $N$ is the order of the input tapped delay and $L$ is the order of the feedback tapped delay line.

**Calculation of BPTT derivatives.** Jacobians are calculated according to the BPTT scheme [1, p. 836], [4], [7]. After calculating the output $\tilde{y}(k+1)$, the NARX network is unfolded back through time. The recurrent neural network is presented as an FFNN with many layers, each corresponding to one retrospective time step $k-1, k-2, ..., k-h$, where $h$ is a BPTT truncation depth. The set of static Jacobians $\frac{\partial \tilde{y}(k-n)}{\partial w}$ are calculated for each of the unrolled retrospective time steps. Finally, dynamic BPTT Jacobians $\frac{\partial \tilde{y}_{BPTT}(k)}{\partial w}$ are averaged static derivatives obtained for the feed-forward layers.

**Extended Kalman Filter method for training NARX.** Training the NARXs using an EKF algorithm is accomplished in the same way as training the DMLPs described above. The only difference is that the observation matrix $H$ is filled by the dynamic derivatives $\frac{\partial \tilde{y}_{BPTT}(k)}{\partial w^{(1)}}$ and $\frac{\partial \tilde{y}_{BPTT}(k)}{\partial w^{(2)}}$, which contain temporal information.

## 2.3 Direct method of training Multilayer Perceptrons for MS predictions using FPTT and Batch EKF

**Calculation of FPTT derivatives.** We propose a new batch-like method of calculating the dynamic derivatives for the FFNNs called Forecasted Propagation Through Time (Fig. 3).

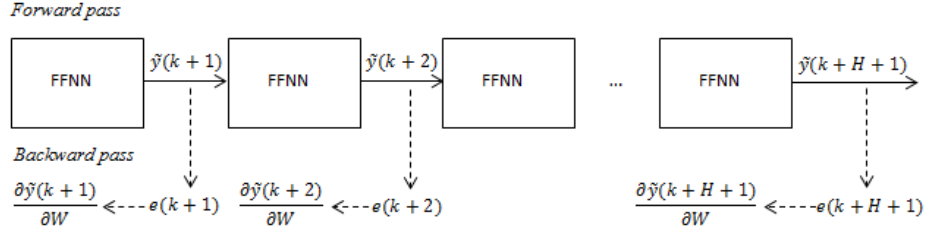

**Fig. 2.** Calculation of dynamic FPTT derivatives for feedforward neural networks

1) At each time step the neural network is unfolded forward through time $H$ times using Eqs. (2)-(3) in the same way as it is performed for regular multi-step-ahead prediction, where $H$ is a horizon of prediction. Outputs $\tilde{y}(k+1),..., \tilde{y}(k+H+1)$ are calculated.
2) For each of $H$ forecasted time steps, prediction errors $e(k+h+1) = t(k+h+1) - \tilde{y}(k+h+1)$, $h = 1,..., H$ are calculated.
3) The set of independent derivatives $\left\{\dfrac{\partial \tilde{y}(k+h)}{\partial W}\right\}$, $h = 1,..., H+1$, using the standard backpropagation of independent errors $\{e(k+h)\}$ are calculated for each copy of the unfolded neural network.

There are three main differences between the proposed FPTT and traditional BPTT. First, BPTT unfolds the neural network backward through time; FPTT unfolds the neural network recursively forward through time. This is useful from a technological point of view because this functionality must be implemented for MS predictions anyway. Second, FPTT does not backpropagate the accumulated error through the whole unfolded structure. It instead calculates BP for each copy of the neural network. Finally, FPTT does not average derivatives, it calculates a set of formally independent errors and a set of formally independent derivatives for future time steps instead. By doing this, we leave the question about contributions of each time step to the total MS error to the Batch Kalman Filter Algorithm.

**Batch Extended Kalman Filter method for training DMLP using FPTT.** The EKF training algorithm also has a batch form [11]. In this case, a batch size of $H$ patterns and a neural network with $L_w$ outputs is treated as training a single shared-weight network with $L_w \times H$ outputs, i.e. $H$ data streams which feed $H$ networks constrained to have identical weights are formed from the training set. A single weight update is calculated and applied equally to each stream's network. This weights update is sub-optimal for all samples in the batch. If streams are taken from different places in the dataset, then this trick becomes equivalent to a Multistream EKF [7], [10], a well-known technique for avoiding poor local minima. However, we use it for direct minimization of accumulated error $H$ steps ahead. Batch observation matrix $\tilde{H}(k)$ and residual matrix $\tilde{E}(k)$ now becomes:

$$\tilde{H}(k) = \begin{bmatrix} \dfrac{\partial \tilde{y}(k+1)}{\partial w_1} & \dfrac{\partial \tilde{y}(k+1)}{\partial w_2} & ... & \dfrac{\partial \tilde{y}(k+1)}{\partial w_{N_w}} \\ ... & ... & ... & ... \\ \dfrac{\partial \tilde{y}(k+H+1)}{\partial w_1} & \dfrac{\partial \tilde{y}(k+H+1)}{\partial w_2} & ... & \dfrac{\partial \tilde{y}(k+H+1)}{\partial w_{N_w}} \end{bmatrix}, \quad (9)$$

$$\tilde{E}(k) = \begin{bmatrix} e(k+1) & e(k+2) & ... & e(k+H+1) \end{bmatrix} \quad (10)$$

The size of matrix $\tilde{R}$ is $(L_w \times H) \times (L_w \times H)$, the size of matrix $\tilde{H}(k)$ is $(L_w \times H) \times N_w$, and the size of matrix $\tilde{E}(k)$ is $(L_w \times H) \times 1$. The remainder is identical to regular EKF.

## 3 Experiments

### 3.1 Mackey-Glass Chaotic Process

The Mackey-Glass chaotic process is a famous benchmark for time series predictions. The discrete-time equation is given by the following difference equation (with delays):

$$x_{t+1} = (1-b)x_t + a \frac{x_{t-\tau}}{1+(x_{t-\tau})^{10}}, t = \tau, \tau+1, ... \quad , \quad (11)$$

where $\tau \geq 1$ is an integer. We used the following parameters: $a = 0.1$, $b = 0.2$, $\tau = 17$ as in [4]. 500 values were used for training; the next 100 values were used for testing.

First, we trained 100 DMLP networks with one hidden layer and hyperbolic tangent activation functions using traditional EKF and BP derivatives. The training parameters for EKF were set as $\eta = 10^{-3}$ and $\mu = 10^{-8}$. The number of neurons in the hidden layer was varied from 3 to 8, the order of input tapped delay line was set $N = 5$, and the initial weights were set to small random values. Each network was trained for 50 epochs. After each epoch, MS prediction on horizon $H = 14$ on training data was performed to select the best network. This network was then evaluated on the test sequence to achieve the final MS quality result. Second, we trained 100 DMLP networks with the same initial weights using the proposed Batch EKF technique together with FPTT derivatives and evaluated their MS prediction accuracy. Third, we trained 100 NARX networks (orders of tapped delay lines: $N = 5, L = 5$) using EKF and BPTT derivatives to make comparisons. The results of these experiments are presented in Table 1. Normalized Mean Square Error (NMSE) was used for the quality estimations.

**Table 1.** Mackey-Glass problem: mean NMSE errors for different prediction horizon values H

|  | H=1 | H=2 | H=6 | H=8 | H=10 | H=12 | H=14 |
|---|---|---|---|---|---|---|---|
| DMLP EKF BP | 0.0006 | 0.0014 | 0.013 | 0.022 | 0.033 | 0.044 | 0.052 |
| DMLP BEKF FPTT | 0.0017 | 0.0022 | 0.012 | 0.018 | 0.022 | 0.027 | 0.030 |
| NARX EKF | 0.0010 | 0.0014 | 0.012 | 0.018 | 0.023 | 0.028 | 0.032 |

### 3.2 Santa-Fe Laser Data Series

In order to explore the capability of the global behavior of DMLP using the proposed training method, we tested it on the laser data from the Santa Fe competition. The dataset consisted of laser intensity collected from the real experiment. Data was divided to training (1000 values) and testing (100 values) subsequences. This time the goal for training was to perform long-term (H=100) MS prediction. The order of the time delay line was set to $N = 25$ as in [2], the rest was the same as in the previous experiment. The obtained average NMSE for 100 DMLP networks was 0.175 for DMLP EKF BP (classic method) versus 0.082 for DMLP BEKF FPTT (proposed method).

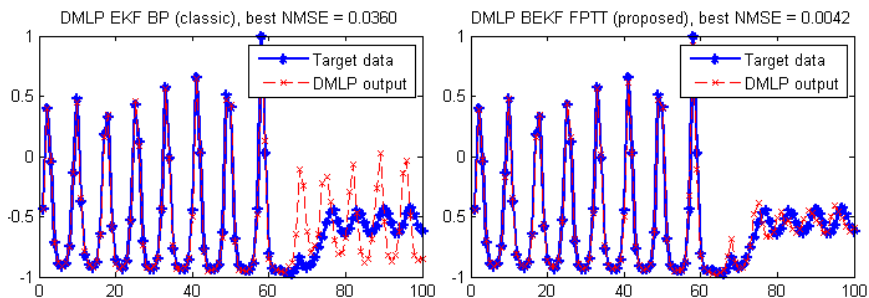

**Fig. 3.** The best results of the closed-loop long-term predictions (H=100) on testing data using DMLPs trained using different methods.

Meanwhile, the best instance trained using Batch EKF+FPTT shows 10 times better accuracy than the best instance trained using the classic approach.

## 4 Conclusions

We considered the multi-step-ahead prediction problem and discussed neural network based approaches as a tool for its solution. Feed-forward and recurrent neural models were considered, and advantages and disadvantages of their usage were discussed. A novel direct method for training feed-forward neural networks to perform multi-step-ahead predictions was proposed, based on the Batch Extended Kalman Filter. This method is considered to be useful from a technological point of view because it uses existing multi-step-ahead prediction functionality for calculating special FPTT dynamic derivatives which require a slight modification of the standard EKF algorithm. Our method demonstrates doubled long-term accuracy in comparison to standard training of the dynamic MLPs using the Extended Kalman Filter due to direct minimization of the accumulated multi-step-ahead error.